\relax
\documentclass[letterpaper]{article} % DO NOT CHANGE THIS
\usepackage{arxiv}  % DO NOT CHANGE THIS
\usepackage{times}  % DO NOT CHANGE THIS
\usepackage{helvet} % DO NOT CHANGE THIS
\usepackage{courier}  % DO NOT CHANGE THIS
\usepackage[hyphens]{url}  % DO NOT CHANGE THIS
\usepackage{graphicx} % DO NOT CHANGE THIS
\usepackage{natbib}  % DO NOT CHANGE THIS AND DO NOT ADD ANY OPTIONS TO IT
\usepackage{caption} % DO NOT CHANGE THIS AND DO NOT ADD ANY OPTIONS TO IT
\usepackage{authblk}
\usepackage{amssymb}
\usepackage{mathtools}
\usepackage{amsthm}
\usepackage{amsmath}
\usepackage{mathpazo}
\usepackage{mathcomp}
\usepackage{stackengine}
\usepackage{extarrows}
\usepackage{array}
\usepackage{algorithmic}
\usepackage{subcaption}
\usepackage{bm}
\usepackage{pseudocode}
\usepackage[ruled, vlined]{algorithm2e}
\usepackage[switch]{lineno}

\DeclarePairedDelimiter\norm{\lVert}{\rVert}%
\DeclarePairedDelimiter\abs{\lvert}{\rvert}%
\DeclareMathOperator*{\argmax}{argmax}

\def\Tiny{\fontsize{6pt}{6pt}\selectfont}
\makeatletter
\def\blfootnote{\xdef\@thefnmark{}\@footnotetext}
\def\delequal{\mathrel{\ensurestackMath{\stackon[1pt]{=}{\scriptstyle\Delta}}}}
\newcommand*{\eqdef}{\ensuremath{\overset{\mathclap{\text{\Tiny def}}}{=}}}

\newcommand{\E}{\mathbb{E}}
\newcommand{\Var}{\mathrm{Var}}

\newtheorem{theorem}{Theorem}[section]
\newtheorem{lemma}[theorem]{Lemma}

\title{OPAC: Opportunistic Actor-Critic}

\author{Srinjoy Roy}
\author{Saptam Bakshi}
\author{Tamal Maharaj}
\affil{	Department of Computer Science\\
	Ramakrishna Mission Vivekananda Educational and Research Institute (RKMVERI)\\
	West Bengal, India\\}

\date{}
\begin{document}
  %  \linenumbers
    \maketitle
	\begin{abstract}
	Actor-critic methods, a type of model-free reinforcement learning (RL), have achieved state-of-the-art performances in many real-world domains in continuous control. Despite their success, the wide-scale deployment of these models is still a far cry. The main problems in these actor-critic methods are inefficient exploration and sub-optimal policies. Soft Actor-Critic (SAC) and Twin Delayed Deep Deterministic Policy Gradient (TD3), two cutting edge such algorithms, suffer from these issues. SAC effectively addressed the problems of sample complexity and convergence brittleness to hyper-parameters and thus outperformed all state-of-the-art algorithms including TD3 in harder tasks, whereas TD3  produced moderate results in all environments. SAC suffers from inefficient exploration owing to the Gaussian nature of its policy which causes borderline performance in simpler tasks. In this paper, we introduce Opportunistic Actor-Critic (OPAC), a novel model-free deep RL algorithm that employs better exploration policy and lesser variance. OPAC combines some of the most powerful features of TD3 and SAC and aims to optimize a stochastic policy in an off-policy way. For calculating the target Q-values, instead of two critics, OPAC uses three critics and based on the environment complexity, opportunistically chooses how the target Q-value is computed from the critics' evaluation. We have systematically evaluated the algorithm on MuJoCo environments where it achieves state-of-the-art performance and outperforms or at least equals the performance of TD3 and SAC.
	\end{abstract}

	\section{Introduction}
	Model-free deep reinforcement learning (RL) algorithms have been successfully applied to a series of challenging domains ranging from games~\cite{DBLP:journals/corr/MnihKSGAWR13, 44806} to robotic control~\cite{DBLP:journals/corr/GuHLL16, DBLP:journals/corr/abs-1803-06773}. The combination of reinforcement learning with powerful function approximators, like neural networks, has given rise to deep reinforcement learning. In recent years, deep RL has proved to be highly effective in a wide range of decision making and control tasks. However, the application of model-free deep RL in such tasks is made complicated by two major challenges -- sample complexity and convergence brittleness. Cutting edge deep RL algorithms like Twin Delayed Deep Deterministic Policy Gradient (TD3)~\cite{DBLP:journals/corr/abs-1802-09477} and Soft Actor-Critic (SAC)~\cite{DBLP:journals/corr/abs-1812-05905} have shown promising results in dealing with challenging control tasks. TD3 learns efficiently from past samples using experience replay memory and it effectively addresses the overestimation bias that occurs in traditional actor-critic methods. But it suffers from the sensitivity to hyper-parameters and as a result, requires a lot of tuning to converge. To combat this convergence brittleness of TD3, the maximum entropy RL framework~\cite{conf/aaai/ZiebartMBD08} was incorporated in SAC.
	
	The reason for the failure of Deep Deterministic Policy Gradient (DDPG)~\cite{journals/corr/LillicrapHPHETS15} based algorithms is the dramatic overestimation of Q-values~\cite{DBLP:journals/corr/abs-1802-09477}. TD3 addressed this issue by the use of three techniques in its algorithm -- clipped double Q-learning, delayed policy updates, and target policy smoothing. It is an off-policy, Q-learning based algorithm which trains a deterministic policy. On the other hand, SAC trains a stochastic policy and explores in an on-policy way. The gap between DDPG style approaches and stochastic policy optimization was bridged by SAC.
	
	The use of target networks has been illustrated in the TD3 algorithm where its role in stabilizing the training process is evident. TD3 follows a pessimistic approach while evaluating the deterministic policy. This is done by the clipped double Q-learning technique where it takes the minimum of two target Q-values for updating the parameters of the critic models. Policy updates and the update of target network parameters are done less frequently than the update of model network parameters. This is to ensure that the error in value network is minimized up to a certain extent before introducing a policy update. Exploration is facilitated in TD3 by adding noise to the target policy to avoid over estimation. SAC also employs the use of target networks but only for the critics. Since it explores in an on-policy way, SAC does not use target networks for the actor, which is the policy itself. The inherent stochastic nature of the policy enables exploration in SAC. It’s analogous to the target policy smoothing in TD3. Entropy regularization is one of the key features of SAC where the policy is trained to maximize a trade-off between the expected return over time and the entropy. The term entropy in this context refers to a measure of randomness in the policy. As already mentioned, TD3 and SAC both employ a pessimistic approach while calculating the Mean Squared Bellman Error (MSBE) by taking the minimum of two Q-values. Optimistic Actor-Critic (OAC) ~\cite{oac}, another recent algorithm, takes an optimistic approach instead. It was shown to attain substantial improvement in the quality of exploration being made.  
	
	In this context, we introduce Opportunistic Actor-Critic (OPAC), a model-free Deep RL algorithm that has incorporated some of the novel features of TD3 and SAC like the use of target networks, target policy smoothing, entropy maximization framework~\cite{conf/aaai/ZiebartMBD08, DBLP:journals/corr/abs-1803-06773, 4739438, 10.1145/1553374.1553508, 10.5555/2540128.2540576}. To introduce the idea of voting, an additional critic is used to fine-tune the value updates. The driving idea behind the development of OPAC is to retain the benefits of TD3 and SAC and combine them under a single roof along with an extra critic to form a link between stochastic policy optimization and off-policy exploration. We demonstrated via experimental results that having three critics instead of two improves the quality of the policy which in turn, yields a higher average reward over time. SAC was shown to outperform TD3 and other model-free Deep RL algorithms (like Proximal Policy Optimization (PPO)~\cite{DBLP:journals/corr/SchulmanWDRK17}, Trust Region Policy Optimization (TRPO)~\cite{DBLP:journals/corr/SchulmanLMJA15}) in challenging tasks like the “Humanoid-v2” environment in MuJoCo. In this paper, we have shown that OPAC outperforms both TD3 and SAC with a few exceptions where it works as par with TD3 and SAC, both in challenging as well as simple control tasks in terms of the average return. Since TD3 and SAC are currently two of the best model-free Deep RL algorithms, we limit our comparison of the performance of OPAC with only TD3 and SAC.

	\section{Background}
	We first discuss the principle concepts regarding reinforcement learning and maximum entropy reinforcement learning. These discussions will contain the necessary mathematical notations that will be useful as well as heavily referred to in the later sections.
	\subsection{Reinforcement Learning}
	Markov Decision Processes (MDPs) are defined by the tuple $(\mathcal{S}, \mathcal{A}, p, r)$, where $\mathcal{S}$ is the finite state space, $\mathcal{A}$ is the finite action space, $p$ represents the state transition probabilities and $r$ represents the reward function. $\mathcal{S}$ and $\mathcal{A}$ are assumed to be continuous and the state transition probability $p : \mathcal{S} \times \mathcal{S} \times \mathcal{A} \rightarrow [0, \infty)$ represents the probability density of the next state $s_{t+1} \in \mathcal{S}$ given the current state $s_{t} \in \mathcal{S}$ and action $a_{t} \in \mathcal{A}$. The goal in an MDP is to find an optimal ``policy'' for the decision maker : a function $\pi$ that specifies the action $\pi(s)$ that the decision maker chooses when in state $s$.
	
	Reinforcement learning (RL) considers the paradigm of an agent interacting with its environment to learn reward-maximizing behavior. The agent in RL could be thought of as the decision-maker in MDPs and the environment could be thought of as the setting on which the MDP is defined. Thus, a standard reinforcement learning framework is defined as a policy search in an MDP. The standard reinforcement learning objective is the expected sum of rewards given by,
	\[\sum_{t = 0}^{\infty} \mathbb{E}_{(s_{t}, a_{t}) \sim \rho_{\pi}}[r(s_{t}, a_{t})].\]
	The goal is to learn a policy $\pi(a_{t} \mid s_{t})$ that maximizes the objective. In other words, we are trying to learn the optimal policy $\pi_{\theta}^{*}(a_{t} \mid s_{t})$, with the parameters $\theta$.
	
	\subsection{Maximum Entropy Reinforcement Learning}
	The maximum entropy objective~\cite{10.5555/2049078} generalizes the standard RL objective by augmenting it with an entropy term, such that the optimal policy additionally aims to maximize its entropy at each visited state:
	\[\pi^{*} = arg \max_{\pi} \sum_{t = 0}^{\infty} \mathbb{E}_{(s_{t}, a_{t}) \sim \rho_{\pi}}[r(s_{t}, a_{t}) + \alpha.\mathcal{H}(\pi(\cdot \mid s_{t}))],\]
	where $\alpha$ is the temperature parameter that determines the relative importance of the entropy term versus the reward, and thus controls the stochasticity of the optimal policy. Entropy is the measure of unpredictability of a random variable. Let $x$ be a random variable with probability mass or density function $\mathcal{P}$. The entropy $\mathcal{H}$ of $x$ is computed from its distribution $\mathcal{P}$ according to,
	\[\mathcal{H}(\mathcal{P}) = \mathbb{E}_{x \sim \mathcal{P}}[- \log \mathcal{P}(x)].\]
	
	The maximum entropy framework has many conceptual and practical advantages. Firstly, the policy is given an incentive to explore more widely, while rejecting actions that are sub-optimal. Secondly, the policy can capture multiple modes of near-optimal behavior and in scenarios where more than one actions seem equally good, the policy will assign equal probabilities to those actions. It has been observed that it considerably improves learning speed over state-of-the-art methods that optimize the standard RL objective function.
	
	\section{Soft Policy Iteration}
	The soft policy iteration is a general algorithm for determining optimal policies under the maximum entropy framework. It alternates between policy improvement and policy evaluation steps. It was introduced and fully derived in the paper of SAC~\cite{DBLP:journals/corr/abs-1812-05905}. We revisit the lemmas and the soft policy iteration theorem but we skip their proofs since those can be found in the aforementioned paper.
	
	Soft policy iteration was shown to converge to an optimal policy within a set of policies. In the policy evaluation step, value of the policy $\pi$ was computed according the maximum entropy reinforcement learning objective. $Q : \mathcal{S} \times \mathcal{A} \rightarrow \mathbb{R}$ was the soft Q-value function~\cite{DBLP:journals/corr/SchulmanAC17,DBLP:journals/corr/NachumNXS17} whose value could be computed iteratively. This was done by repeatedly applying a modified Bellman backup operator $\mathcal{T}^{\pi}$ defined by,
	\begin{equation}\label{eqn:Equation1}
		\mathcal{T}^{\pi} Q(s_{t}, a_{t}) \delequal r(s_{t}, a_{t}) + \gamma \mathbb{E}_{s_{t + 1} \sim p}[V(s_{t + 1})] ,
	\end{equation}
	where,
	\begin{equation}\label{eqn:Equation2}
		V(s_{t}) = \mathbb{E}_{a_{t} \sim \pi}[Q(s_{t}, a_{t}) - \alpha \log \pi(a_{t} \mid s_{t})]
	\end{equation}
	was the soft state value function. The soft Q-function for any policy $\pi$ was obtained by repeatedly applying $\mathcal{T}^{\pi}$ as was formalized in Lemma \ref{th:lemma0.1}.
	\begin{lemma}[Soft Policy Evaluation]\label{th:lemma0.1}
		Consider the soft Bellman backup operator $\mathcal{T}^{\pi}$ in Equation \ref{eqn:Equation1} and a mapping $Q^{0} : \mathcal{S} \times \mathcal{A} \rightarrow \mathbb{R}$ with $|\mathcal{A}| < \infty$ and define $Q^{k + 1} = \mathcal{T}^{\pi} Q^{k}$. Then the sequence $Q^{k}$ will converge to the soft Q-function of $\pi$ as $k \rightarrow \infty$.
	\end{lemma}
	In the policy improvement step, for each state, the policy was updated according to,
	\begin{equation}\label{eqn:Equation3}
	\pi_{new} = arg \min_{\pi^{'} \in \Pi} D_{KL} \left(\pi^{'}(\cdot \mid s_{t}) \left| \right| \frac{exp(\frac{1}{\alpha} Q^{\pi_{old}}(s_{t}, \cdot))}{Z^{\pi_{old}}(s_{t})} \right),
	\end{equation}
	where $\pi_{new}$ corresponded to the updated policy and was updated towards the exponential of the new soft Q-function. $\Pi$ was a set of policies which belonged to the parameterized family of Gaussian distributions. Information projection defined in terms of the Kullback-Leibler divergence was used to project the improved policy into the desired set of policies to satisfy the constraint $\pi^{'} \in \Pi$. $Z^{\pi_{old}}(s_{t})$ was the partition function which normalized the distribution. It was ignored because it did not contribute to the gradient with respect to $\pi_{new}$. In Lemma \ref{th:lemma0.2} it was formalized that the new projected policy had a higher value than the old policy.
	\begin{lemma}[Soft Policy Improvement]\label{th:lemma0.2}
		Let $\pi_{old} \in \Pi$ and let $\pi_{new}$ be the optimizer of the minimization problem defined in Equation \ref{eqn:Equation3}. Then $Q^{\pi_{new}}(s_{t}, a_{t}) \geq Q^{\pi_{old}}(s_{t}, a_{t})$ for all $(s_{t}, a_{t}) \in \mathcal{S} \times \mathcal{A}$ with $|\mathcal{A}| < \infty$.
	\end{lemma}
	In Theorem~\ref{th:theorem0.3} it was proved that the soft policy iteration algorithm converges to the optimal maximum entropy policy by alternating between soft policy evaluation and soft policy improvement steps.
	\begin{theorem}[Soft Policy Iteration]\label{th:theorem0.3}
	    Repeated application of soft policy evaluation and soft policy improvement from any $\pi \in \Pi$ converges to a policy $\pi^{*}$ such that $Q^{\pi^{*}}(s_{t}, a_{t}) \geq Q^{\pi}(s_{t}, a_{t})$ for all $\pi \in \Pi$ and $(s_{t}, a_{t}) \in \mathcal{S} \times \mathcal{A}$ with $|\mathcal{A}| < \infty$.
	\end{theorem}

	\section{Opportunistic Actor-Critic}
	Soft policy iteration was derived in a tabular setting. To extend this into continuous state-action domains the soft Q-function and the policy both, have to be approximated by the use of deep neural networks. Instead of alternating between soft policy evaluation and soft policy improvement up to convergence, we will alternate between optimizing the soft Q-function and policy network by stochastic gradient descent. This is how we will construct our algorithm of OPAC. Let $Q_{\phi}(s_{t}, a_{t})$ and $\pi_{\theta}(a_{t} \mid s_{t})$ be the soft Q-function and a tractable policy with parameters $\phi$ and $\theta$ respectively.
	Parameters of the soft Q-function can be trained to minimize the soft Bellman residual error,
	\begin{align*}
		J_{Q}(\phi) &= \mathbb{E}_{(s_{t}, a_{t}) \sim D}[\frac{1}{2}(Q_{\phi}(s_{t}, a_{t}) - (r(s_{t}, a_{t}) 
		+ \gamma \mathbb{E}_{s_{t + 1} \sim p}[V_{\phi_{target}}(s_{t + 1})]))^{2}]
	\end{align*}
	Substituting the value function parameters as in Equation~\ref{eqn:Equation2} in the above equation and optimizing it by stochastic gradient descent we have,
	\begin{align}\label{eqn:Equation4}
		\nabla&_{\phi}J_{Q}(\phi) = \nabla_{\phi}Q_{\phi}(s_{t}, a_{t})(Q_{\phi}(s_{t}, a_{t}) - (r(s_{t}, a_{t}) + \gamma Q_{\phi_{target}}(s_{t + 1}, a_{t + 1}) - \alpha \log \pi_{\theta_{target}}(a_{t + 1} \mid s_{t + 1}))),
	\end{align}
	where $\phi_{target}$ in the update rule denotes the parameters of the target Q-function networks. This is an important tool for stabilizing training~\cite{mnih2015humanlevel}. The parameters of the policy network can be directly learned by minimizing the KL-divergence in Equation \ref{eqn:Equation3},
	\begin{equation}\label{eqn:Equation5}
		J_{\pi}(\theta) = \mathbb{E}_{s_{t} \sim D}[\mathbb{E}_{a_{t} \sim \pi_{\theta}}[\alpha \log \pi_{\theta}(a_{t} \mid s_{t}) - Q_{\phi}(s_{t}, a_{t})]].
	\end{equation}
	We need to compute,
	\[\nabla_{\theta}\mathbb{E}_{a_{t} \sim \pi_{\theta}}[\alpha \log \pi_{\theta}(a_{t} \mid s_{t}) - Q_{\phi}(s_{t}, a_{t})].\]
	$Q_{\phi}(s_{t}, a_{t})$ does not directly depend on $\theta$, thus no gradient of $Q_{\phi}(s_{t}, a_{t})$ can be computed over $\theta$. Rather, we can write the action $a_{t}$ as,
	\[a_{t} = \mu_{\theta}(s_{t}) + \epsilon_{t} \sigma_{\theta}(s_{t}),\]
	where $\epsilon_{t} \sim \mathcal{N}(0, 1)$. Instead of sampling $a_{t} \sim \pi_{\theta}(s_{t})$, we now sample $\epsilon_{t} \sim \mathcal{N}(0, 1)$. Therefore, we can surely write
	\[Q_{\phi}(s_{t}, a_{t}) = Q_{\phi}(s_{t}, \mu_{\theta}(s_{t}) + \epsilon_{t} \sigma_{\theta}(s_{t})).\]
	Thus, a gradient over $\theta$ appears, leading to smaller variance. We set $a_{t} = f_{\theta}(\epsilon_{t}, s_{t}) = \mu_{\theta}(s_{t}) + \epsilon_{t} \sigma_{\theta}(s_{t})$. Now we have,
	\begin{align*}
		J_{\pi}(\theta) &= \mathbb{E}_{s_{t} \sim D}[\mathbb{E}_{\epsilon_{t} \sim \mathcal{N}(0, 1)}[\alpha \log \pi_{\theta}(f_{\theta}(\epsilon_{t}, s_{t}) \mid s_{t}) - Q_{\phi}(s_{t}, f_{\theta}(\epsilon_{t}, s_{t}))]]
	\end{align*}
	whose gradient with respect to $\theta$ can be obtained by,
	\begin{align*}
		\nabla_{\theta} J_{\pi}(\theta) &= \nabla_{\theta} \log \pi_{\theta}(a_{t} \mid s_{t}) + (\nabla_{a_{t}} \log \pi_{\theta}(a_{t} \mid s_{t}) - \nabla_{a_{t}} Q_{\phi}(s_{t}, a_{t}))\nabla_{\theta}f_{\theta}(\epsilon_{t}, s_{t}).
	\end{align*}
	Finally, we have all the necessary update rules for OPAC. The whole process described in this section has a lot of similarity with that of SAC especially in the use of reparameterization trick. However, in practice it has been observed that learning policy parameters by the above equation yields inferior results. Instead, we can learn the policy parameters by,
	\begin{equation}\label{eqn:Equation6}
		J_{\pi}(\theta) = \mathbb{E}_{s_{t} \sim D}[\mathbb{E}_{a_{t} \sim \pi_{\theta}}[\alpha \log \pi_{\theta}(a_{t} \mid s_{t}) - Q_{\phi_{1}}(s_{t}, a_{t})]]
	\end{equation}
	Note that $Q_{\phi}(s_{t}, a_{t})$ has become $Q_{\phi_{1}}(s_{t}, a_{t})$. The significance of $\phi_{1}$ is that we are only considering the output of the first Q-network. The $J_{\pi}(\theta)$ in Equation \ref{eqn:Equation6} can be optimized by stochastic gradient descent using a similar reparameterization trick as of Equation \ref{eqn:Equation5}. This modification was inspired from the policy update rule of TD3. Main reason for modifying policy update rule of Equation \ref{eqn:Equation5} to what's in Equation \ref{eqn:Equation6} is strictly for practical purposes. We will look at it more deeply in an upcoming section where we present an algorithm for OPAC.
	
	\section{Automatic Entropy Adjustment}
	In the previous section, we constructed an off-policy algorithm for OPAC given a particular temperature i.e., the value of $\alpha$ was fixed. Figuring out an optimal temperature is, in practice, a complicated task. The entropy can vary unpredictably both across tasks and during training as the policy becomes better. We borrow the same strategy that SAC uses to automatically adjust the entropy temperature $\alpha$.
	
	The standard maximum entropy learning problem for OPAC can be reformulated as a constraint optimization problem - while maximizing the expected return, the policy should satisfy a minimum entropy constraint, $\max_{\pi_{0}...\pi_{T}} \mathbb{E}[\sum_{t = 0}^{T}r(s_{t}, a{t})]$ s.t. $\forall t$, $\mathcal{H}(\pi_{t}) \geq \mathcal{H}_{0}$, where $\mathcal{H}_{0}$ is a predefined minimum policy entropy threshold. The expected return $\mathbb{E}[\sum_{t = 0}^{T}r(s_{t}, a{t})]$ can be decomposed into a sum of rewards at all the time steps. We make use of a dynamic programming strategy. Since the policy $\pi_{t}$ at time $t$ has no effect on the policy at the earlier time step $\pi_{t - 1}$, we can maximize the return at different steps backward in time.
	\[\max_{\pi_{0}}(\mathbb{E}[r(s_{0}, a_{0})] + \max_{\pi_{1}}(\mathbb{E}[...] + \max_{\pi_{T}}\mathbb{E}[r(s_{T}, a_{T})])),\]
	where we consider $\gamma = 1$. So we start the optimization from the last timestep $T$:
	\[\text{maximize}(\mathbb{E}_{(s_{T}, a_{T}) \sim \rho_{\pi}}[r(s_{T}, a_{T})])\]
	such that, $\mathcal{H}(\pi_{t}) - \mathcal{H}_{0} \geq 0$. Firstly, let us define the following functions:
	\begin{align*}
	h(\pi_{T}) = \mathcal{H}(\pi_{T}) - \mathcal{H}_{0} = \mathbb{E}_{(s_{T}, a_{T}) \sim \rho_{\pi}}[-\log \pi_{T}(a_{T} \mid s_{T})] - \mathcal{H}_{0}
	\end{align*}
	\begin{equation*}
	f(\pi_{T}) = \begin{cases}
	\mathbb{E}_{(s_{T}, a_{T}) \sim \rho_{\pi}}[r(s_{T}, a_{T})],	&\text{if }h(\pi_{T}) \geq 0\\
	-\infty,														&\text{otherwise}
	\end{cases}
	\end{equation*}
	Then the optimization problem becomes,
	\[\text{maximize}(f(\pi_{T})) \text{ s.t. }h(\pi_{T}) \geq 0.\]
	To solve this maximization optimization with inequality constraint, we can construct a Lagrangian expression with a Lagrange multiplier $\alpha_{T}$ as,
	\[L(\pi_{T}, \alpha_{T}) = f(\pi_{T}) + \alpha_{T}h(\pi_{T}).\]
	We skip rest of the part where we minimize $L(\pi_{T}, \alpha_{T})$ with respect to $\alpha_{T}$ - given a particular value $\pi_{T}$, because a similar approach is already given in~\cite{DBLP:journals/corr/abs-1812-05905}. Therefore, we can conclude that we will have equations of the following form,
	\begin{align*}
		\alpha^{*}_{T - 1} &= \text{arg} \min_{\alpha_{T - 1} \geq 0}\mathbb{E}_{(s_{T - 1}, a_{T - 1}) \sim \rho_{\pi^{*}}}[\alpha_{T - 1}\mathcal{H}(\pi^{*}_{T - 1}) - \alpha_{T - 1}\mathcal{H}_{0}]
	\end{align*}
	and,
	\[\alpha^{*}_{T} = \text{arg} \min_{\alpha_{T} \geq 0}\mathbb{E}_{(s_{T}, a_{T}) \sim \rho_{\pi^{*}}}[\alpha_{T}\mathcal{H}(\pi^{*}_{T}) - \alpha_{T}\mathcal{H}_{0}],\]
	where, $\alpha^{*}_{T}$ corresponds to the optimal temperature at the last timestep $T$. The equation for updating $\alpha^{*}_{T - 1}$ has the same form as the equation for updating $\alpha^{*}_{T}$. By repeating this process, we can learn the optimal temperature parameter in every step by minimizing the objective function:
	\begin{equation}\label{eqn:Equation7}
		J(\alpha) = \mathbb{E}_{a_{t} \sim \pi_{t}}[-\alpha \log \pi_{t}(a_{t} \mid s_{t}) - \alpha\mathcal{H}_{0}].
	\end{equation}
	
	\section{Clipped Triple Q-learning}
	Unlike TD3 and SAC our algorithm of OPAC uses clipped triple Q-learning instead of clipped double Q-learning. But practically, we are considering two strategies -- mean value of the smaller two critics and median value of all the three critics. We now establish a proof of convergence for clipped triple Q-learning. The convergence of the mean and median strategies will automatically follow from this proof.
	
	We first include a lemma due to~\cite{10.1023/A:1007678930559} which we are going to use for the convergence proof of Triple Q-learning. It originally appears as a proposition in~\cite{10.5555/517430} which was further generalised into this lemma. The proof of Triple Q-learning is similar to the proof of double Q-learning~\cite{NIPS2010_3964} and Clipped Double Q-learning~\cite{DBLP:journals/corr/abs-1802-09477}.
	\begin{lemma} \label{th:master_lemma}
		Consider a stochastic process $ (\zeta_{t}, \Delta_{t}, F_{t}), t \geq 0$ where $\zeta_{t}, \Delta_{t}, F_{t} : X \to \mathbb{R} $ satisfy the equation:
		\begin{equation*}
		\Delta_{t + 1}(x_{t}) = (1 - \zeta_{t}(x_{t})) \Delta_{t}(x_{t}) + \zeta_{t}(x_{t}) F_{t}(x_{t}),
		\end{equation*}
		where, $ x_{t} \in X $ and $ t = 0, 1, 2,\dots$. Let $P_{t} $ be a sequence of increasing $ \sigma$-fields such that $\zeta_{0}$ and $\Delta_{0} $ are $P_{0}$ measurable and $\zeta_{t}, \Delta_{t}$ and $F_{t - 1}$ are $P_{t}$ measurable, $ t = 0, 1, 2, \dots $. Assume that the following hold:
		\begin{enumerate}
			\item The set X is finite.
			\item $ \zeta_{t}(x_{t}) \in [0, 1] $, $ \sum_{t} \zeta_{t}(x_{t}) = \infty $, $ \sum_{t} (\zeta_{t}(x_{t}))^2 < \infty $ with probability 1 and $ \forall x \neq x_{t} : \zeta_{t} = 0 $.
			\item $ \norm{\E[F_{t} | P_{t}]} \leq \kappa \norm{\Delta_{t}} + c_{t} $ where $ \kappa \in [0, 1) $ and $ c_{t} $ converges to 0 with probability 1.
			\item $ \Var{[F_{t} | P_{t}]} \leq K(1 + \kappa \norm{\Delta_{t}})^2 $, where $ K $ is some constant.
		\end{enumerate}
		Where $\norm{\; . \;}$ denotes the maximum norm. Then $ \Delta_{t} $ converges to 0 with probability 1.
	\end{lemma}
	For a finite MDP setting, we maintain 3 tabular estimates of the value functions $ Q^{A} $, $ Q^{B} $, and $ Q^{C} $. At each timestep we update all of them.
	\begin{theorem}[Clipped Triple Q-learning]
		Given the following conditions:
		\begin{enumerate}
			\item Each state action pair is sampled an infinite number of times.
			\item The MDP is finite.
			\item $ \gamma \in [0, 1) $.
			\item Q-values are stored in a lookup table.
			\item $ Q^{A} $, $ Q^{B} $, and $ Q^{C} $ receive an infinite number of updates.
			\item The learning rates satisfy the following conditions: $ \alpha_{t}(s, a) \in [0, 1] $, $ \sum_{t} \alpha_{t}(s, a) = \infty $, $ \sum_{t} \left(\alpha_{t}(s, a)\right)^2 < \infty $ with probability 1, and $ \alpha_{t}(s, a) = 0 $, $ \forall (s,a) \neq (s_{t}, a_{t}) $.
			\item $ \Var{[r(s, a)]} < \infty, \forall (s, a) $.
		\end{enumerate}
		Then Clipped Triple Q-learning will converge to the optimal action value function $ Q^{*} $, as deﬁned by the Bellman optimality equation, with probability 1.
	\end{theorem}
	\begin{proof}
		We apply lemma \ref{th:master_lemma} with $ P_{t} = \{Q_{0}^{A}, Q_{0}^{B}, Q_{0}^{C}, s_{0}, a_{0}, \alpha_{0}, r_{1}, s_{1}, \dots, s_{t}, a_{t}\} $, $ X = S \times A $, $ \zeta_{t} = \alpha_{t} $. Consider a target mapping, $ g : Q_{t}^{A} \times Q_{t}^{B} \times Q_{t}^{C} \mapsto q $, $ q \in \mathbb{R} $. Also without loss of generality, let's assume $ \Delta_{t} = Q_{t}^{A} - Q^{*} $.
		
		The condition 1 and 4 of lemma \ref{th:master_lemma} holds by the conditions 2 and 4 of the theorem respectively. Lemma condition 2 holds by the theorem condition 6 along with our selection of $ \zeta_{t} = \alpha_{t} $.
		
		Defining $a^{*} = \argmax_{a} Q^{A}(s_{t + 1}, a)$ we have,
		\begin{align*}
		    \Delta_{t + 1}(s_{t}, a_{t}) ={}& (1 - \alpha_{t}(x_{t}))(Q_{t}^{A}(s_{t}, a_{t}) - Q^{*}(s_{t}, a_{t}))\\
		    {}&  + \alpha_{t}(x_{t})(r_{t} + \gamma g \left( Q_{t}^{A}(s_{t + 1}, a^{*}),  Q_{t}^{B}(s_{t + 1}, a^{*}), Q_{t}^{C}(s_{t + 1}, a^{*}) \right)  - Q^{*}(s_{t}, a_{t}))\\
		    ={}& (1 - \alpha_{t}(s_{t}, a_{t})) \Delta_{t}(s_{t}, a_{t}) + \alpha_{t}(s_{t}, a_{t}) F_{t}(s_{t}, a_{t})
	    \end{align*}
	    where, $F_{t}(s_{t}, a_{t})$ is defined as:
	    \begin{align}
		    F_{t}(s_{t}, a_{t}) ={}& r_{t} + \gamma g \left( Q_{t}^{A}(s_{t + 1}, a^{*}), Q_{t}^{B}(s_{t + 1}, a^{*}),  Q_{t}^{C}(s_{t + 1}, a^{*}) \right) - Q^{*}(s_{t}, a_{t}) \nonumber \\
		    ={}& r_{t} + \gamma Q_{t}^{A}(s_{t + 1}, a^{*}) - Q^{*}(s_{t}, a_{t}) + \gamma g \left( Q_{t}^{A}(s_{t + 1}, a^{*}), Q_{t}^{B}(s_{t + 1}, a^{*}), Q_{t}^{C}(s_{t + 1}, a^{*}) \right) - \gamma Q_{t}^{A}(s_{t + 1}, a^{*}) \nonumber \\
		    ={}& F_{t}^{Q}(s_{t}, a_{t}) + c_{t}
	    \end{align}
	    where,
	        \[F_{t}^{Q}(s_{t}, a_{t}) = r_{t} + \gamma Q_{t}^{A}(s_{t + 1}, a^{*}) - Q^{*}(s_{t}, a_{t})\]
	    and,
	        \begin{align*}
		        c_{t} &= \gamma g \left( Q_{t}^{A}(s_{t + 1}, a^{*}), Q_{t}^{B}(s_{t + 1}, a^{*}), Q_{t}^{C}(s_{t + 1}, a^{*}) \right) - \gamma Q_{t}^{A}(s_{t + 1}, a^{*}).
	        \end{align*}
	    $F_{t}^{Q}$ denotes the value of $F_{t}$ under the standard Q-learning. $\E{[F_{t}^{Q} | P_{t}]} \leq \gamma \norm{\Delta_{t}}$ is known to be true due to Bellman operator being a \textit{contraction mapping}. This implies condition 3 of lemma \ref{th:master_lemma} holds if we can show that $c_{t}$ converges to 0 with probability 1. Let, $y = r_{t} + \gamma g \left( Q_{t}^{A}(s_{t + 1}, a^{*}), Q_{t}^{B}(s_{t + 1}, a^{*}), Q_{t}^{C}(s_{t + 1}, a^{*}) \right)$, $\Delta_{t}^{BA} = Q_{t}^{B}(s_{t}, a_{t}) - Q_{t}^{A}(s_{t}, a_{t})$, and $\Delta_{t}^{BC} = Q_{t}^{B}(s_{t}, a_{t}) - Q_{t}^{C}(s_{t}, a_{t})$. It means $c_{t}$ converges to 0 if both $\Delta_{t}^{BA}$ and $\Delta_{t}^{BC}$ converges to 0 with probability 1. Again,
	    \begin{align*}   \label{eq:3}
	        \Delta_{t + 1}^{BA}(s_{t}, a_{t}) \eqdef{}& Q_{t + 1}^{B}(s_{t}, a_{t}) - Q_{t + 1}^{A}(s_{t}, a_{t})\\
	        \begin{split}   \nonumber
	            ={}& \left[Q_{t}^{B}(s_{t}, a_{t}) + \alpha_{t}(s_{t}, a_{t}) \left(y - \; Q_{t}^{B}(s_{t}, a_{t})\right)\right] - \left[Q_{t}^{A}(s_{t}, a_{t})  +  \alpha_{t}(s_{t}, a_{t}) \left(y - Q_{t}^{A}(s_{t}, a_{t})\right)\right]
	        \end{split}\\
	        \begin{split}   \nonumber
	            ={}& \left(Q_{t}^{B}(s_{t}, a_{t}) - Q_{t}^{A}(s_{t}, a_{t})\right)  - \alpha_{t}(s_{t}, a_{t})\left(Q_{t}^{B}(s_{t}, a_{t}) - Q_{t}^{A}(s_{t}, a_{t}\right)
	        \end{split}\\
	        \begin{split}   \nonumber
	            ={}& \Delta_{t}^{BA}(s_{t}, a_{t}) - \alpha_{t}(s_{t}, a_{t}) \Delta_{t}^{BA}(s_{t}, a_{t})
	        \end{split} \\
	        \begin{split}   \nonumber
	            ={}& \left(1 - \alpha_{t}(s_{t}, a_{t})\right) \Delta_{t}^{BA}(s_{t}, a_{t})
	        \end{split}
	    \end{align*}
	
	    \begin{algorithm}
	    \label{alg:Algorithm1}
		\SetAlgoLined
		\KwIn{Initial policy parameters $\theta$, Q-function parameters $\phi_{1}$, $\phi_{2}$, $\phi_{3}$ and an empty replay buffer $D$.}
		Set target parameters equal to main parameters, $\theta_{target} \leftarrow \theta$, $\phi_{target, 1} \leftarrow \phi_{1}$, $\phi_{target, 2} \leftarrow \phi_{2}$, $\phi_{target, 3} \leftarrow \phi_{3}$.\\
		Populate the replay buffer $D$.\\
		\Repeat{convergence}{
		    If $s^{'}$ is a terminal state, reset the environment.\\
		        \If{it's time to update}{
		            \For{j in range(number of updates)}{
		                Sample a batch of transitions, $B = \{(s, a, r, s^{'}, d)\}$ from $D$.
					
					    Compute target actions,\[a^{'}(s^{'}) = clip(\pi_{\theta_{target}}(s^{'}) + clip(\epsilon, -c, c), a_{low}, a_{high}),\text{ where } \epsilon \sim \mathcal{N}(0, \sigma).\] 
					
					    Compute the shared Q-target by,
					    \begin{align*}
					        y(r, s^{'}, d) = r + \gamma(1 - d)(\text{mean/median}
					        - \alpha.\log \pi_{\theta_{target}}(a^{'} \mid s^{'}))
					    \end{align*}
					
					    Update Q-functions (critic-models) using, \[\nabla_{\phi_{i}}.\frac{1}{\abs B} \sum_{(s, a, r, s^{'}, d) \in B}(Q_{\phi_{i}}(s, a) - y(r, s^{'}, d))^{2},\text{ for i = 1, 2, 3}.\]\\
					    \If{ j $\mod$ policy\textunderscore delay = 0}{
						Update policy by,	\[\nabla_{\theta}.\frac{1}{\abs B} \sum_{s \in B}(Q_{\phi_{1}}(s, \pi_{\theta}(s)) - \alpha.\log \pi_{\theta}(\pi_{\theta}(s) \mid s))\]\\
					    Update the target networks and adjust temperature $\alpha$ (for i = 1, 2, 3),
						\begin{center}
							$\theta_{target} \gets \tau.\theta_{target} + (1 - \tau)\theta$\\
							$\phi_{target, i} \gets \tau.\phi_{target, i} + (1 - \tau)\phi_{i}$\\
							$\alpha \gets \alpha - \lambda \nabla_{\alpha}J(\alpha)$
						\end{center}
					}
				}
			}
		}
		\caption{OPAC}
	\end{algorithm}
	
	    Clearly, $\Delta_{t}^{BA}$ converges to 0. Using similar arguments, it can be shown that $\Delta_{t}^{BC}$ converges to 0. These imply we have fulfilled the condition 3 of lemma \ref{th:master_lemma}, implying $Q^{A}(s_{t}, a_{t})$ converges to $Q_{t}^{*}(s_{t}, a_{t})$. Similarly, it can be shown that $Q^{B}(s_{t}, a_{t})$ and $Q^{C}(s_{t}, a_{t})$ converge to the optimal action value function by choosing $\Delta_{t} = Q_{t}^{B} - Q^{*}$ and $\Delta_{t} = Q_{t}^{C} - Q^{*}$ respectively.
	\end{proof}
	
	\section{The Opportunistic Actor-Critic Algorithm}
	The final algorithm for OPAC is listed in Algorithm 1. It makes use of three soft Q-functions, i.e., critics to reduce positive bias in the policy improvement step that is known to degrade the performance of value-based methods \cite{NIPS2010_3964, DBLP:journals/corr/abs-1802-09477}. Since each of the three soft Q-functions have parameters $\phi_{i}$, where $i = 1, 2, 3$, we train them independently to optimize $J_{Q}(\phi_{i})$. We then use two strategies -- the mean value of the smaller two critics and the median value of all the three critics for computing the stochastic gradient in Equation \ref{eqn:Equation4} and policy gradient in Equation \ref{eqn:Equation6}. It is important to note that the policy gradient is computed by gradient ascent once every two iterations while the gradients for the soft Q-functions are computed by stochastic gradient descent in every iteration. Algorithm 1 makes use of two variables ``mean'' and ``median'' which store mean value and the median value respectively, according to the strategies mentioned earlier. The algorithm either uses "mean" or "median" in a single instance.
	
	The entropy temperature $\alpha$ is learned automatically by minimizing the objective function in Equation \ref{eqn:Equation7}. There are target networks for the policy (i.e., the actor) and the three soft Q-functions (i.e., the critics). In short, there are $8$ deep neural networks in our algorithm - one for the actor target and actor model each and, three for critic targets and critic models each. The target networks are updated by Polyak averaging once every two iterations. Gaussian noise is added to the actions $a^{'}$ played by the actor-target for target policy smoothing. The added Gaussian noise can also be termed as exploration noise and it is clipped in the algorithm to keep the target close to the original action.
	
	\section{Experiments}
	We have selected six environments namely Ant-v2, HalfCheetah-v2, Hopper-v2, Humanoid-v2, InvertedPendulum-v2, and Walker2d-v2 for comparing the performance of OPAC with SAC and TD3. All the algorithms have been tested in the MuJoCo continuous control tasks \cite{6386109} interfaced through OpenAI Gym \cite{1606.01540}.
	
	\begin{figure*}[ht!]
		\centering
		\begin{subfigure}[b]{0.27\textwidth}
			\includegraphics[width=\textwidth]{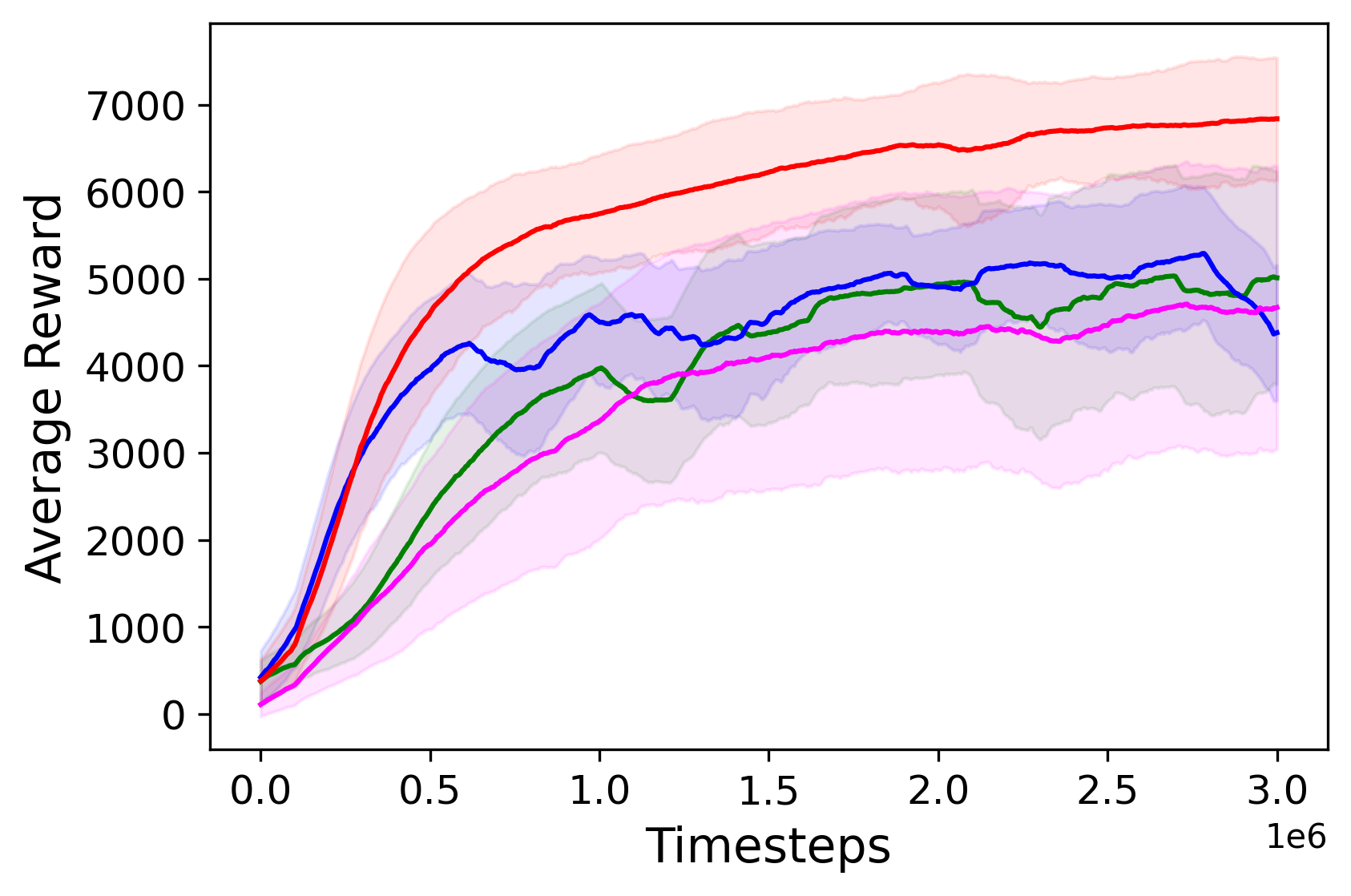}
			\caption{Ant-v2}
		\end{subfigure}
		\begin{subfigure}[b]{0.27\textwidth}
			\includegraphics[width=\textwidth]{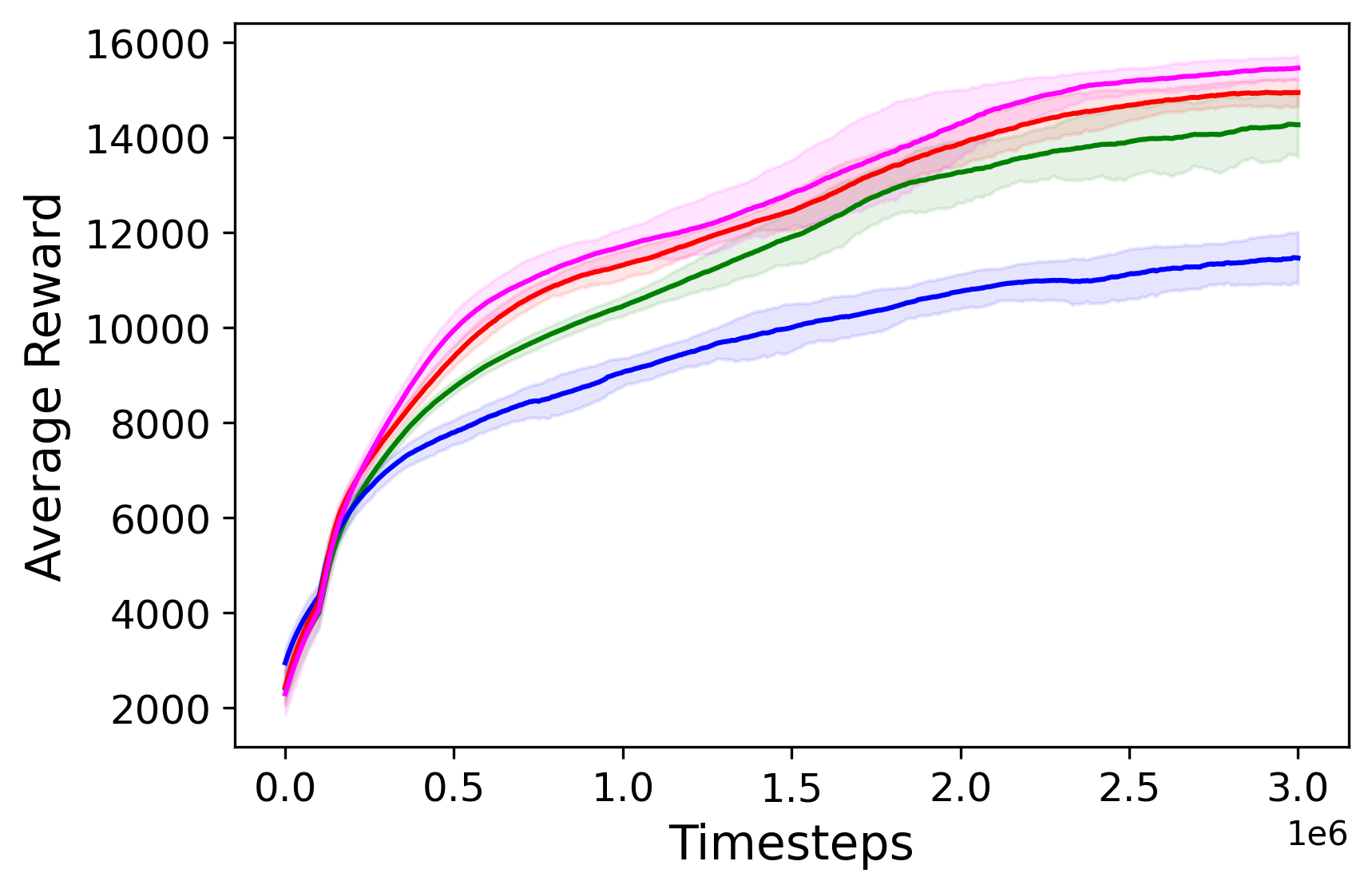}
			\caption{HalfCheetah-v2}
		\end{subfigure}
		\begin{subfigure}[b]{0.27\textwidth}
			\includegraphics[width=\textwidth]{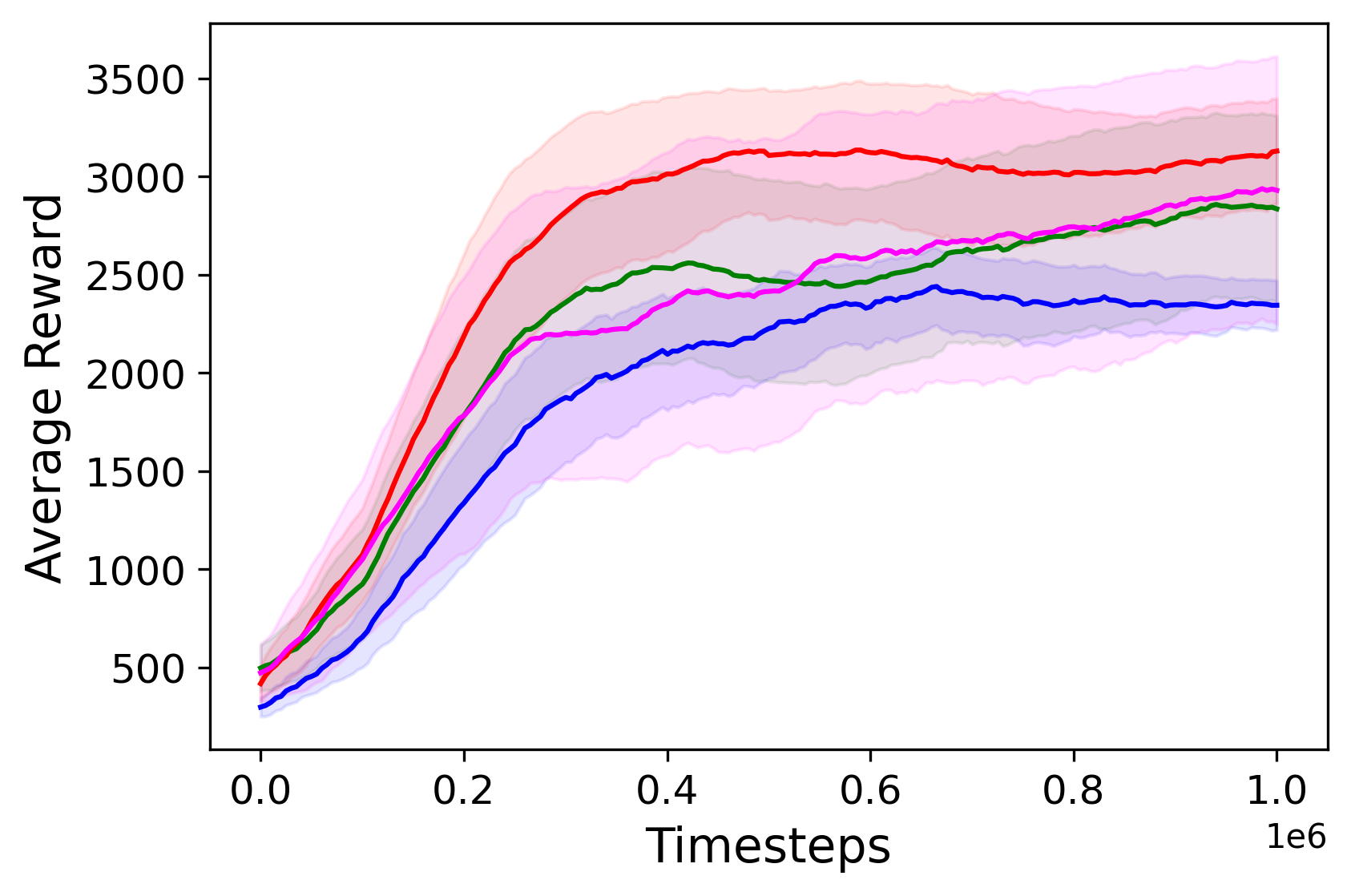}
			\caption{Hopper-v2}
		\end{subfigure}
		\begin{subfigure}[b]{0.27\textwidth}
			\includegraphics[width=\textwidth]{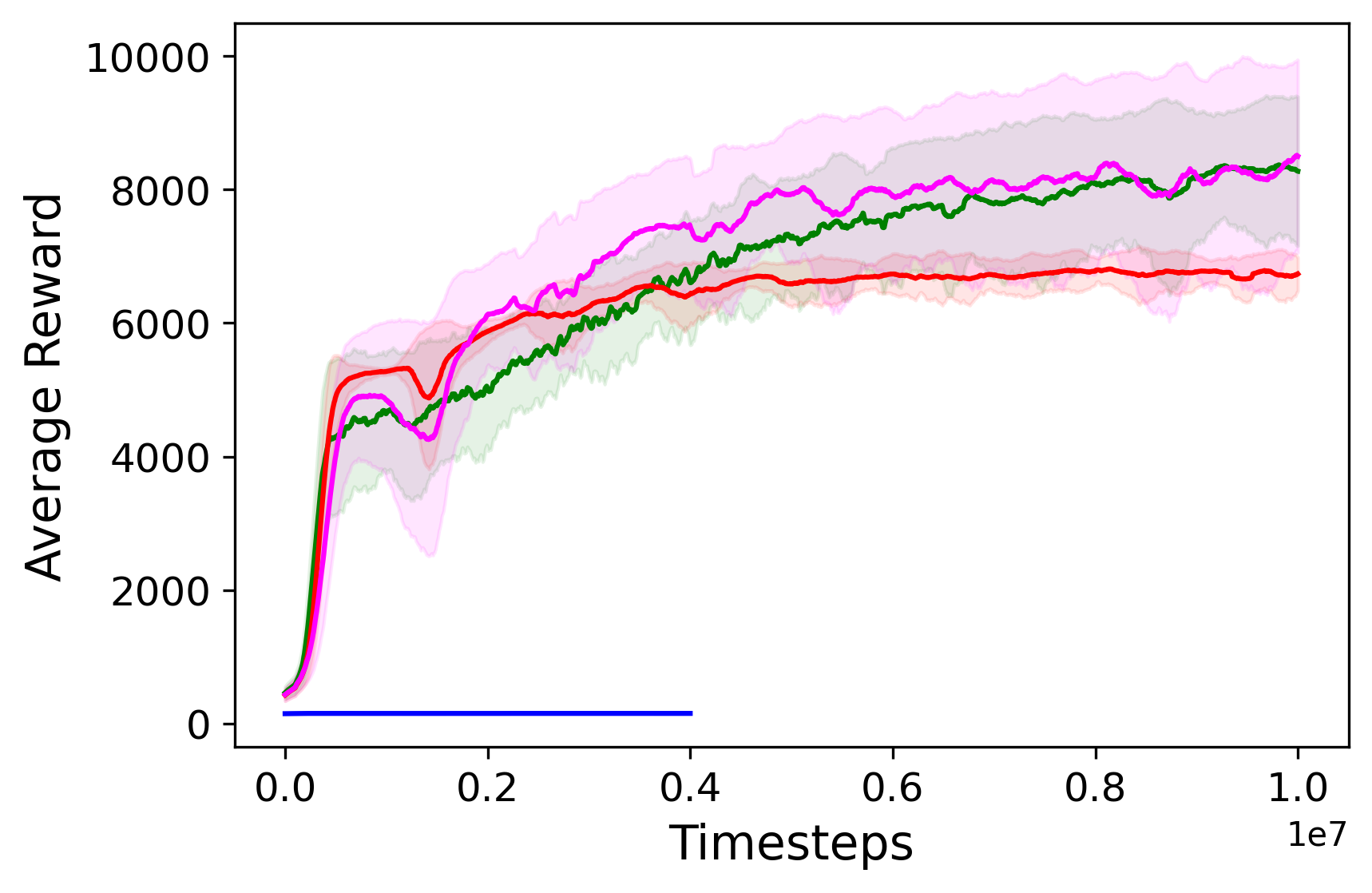}
			\caption{Humanoid-v2}
		\end{subfigure}
		\begin{subfigure}[b]{0.27\textwidth}
			\includegraphics[width=\textwidth]{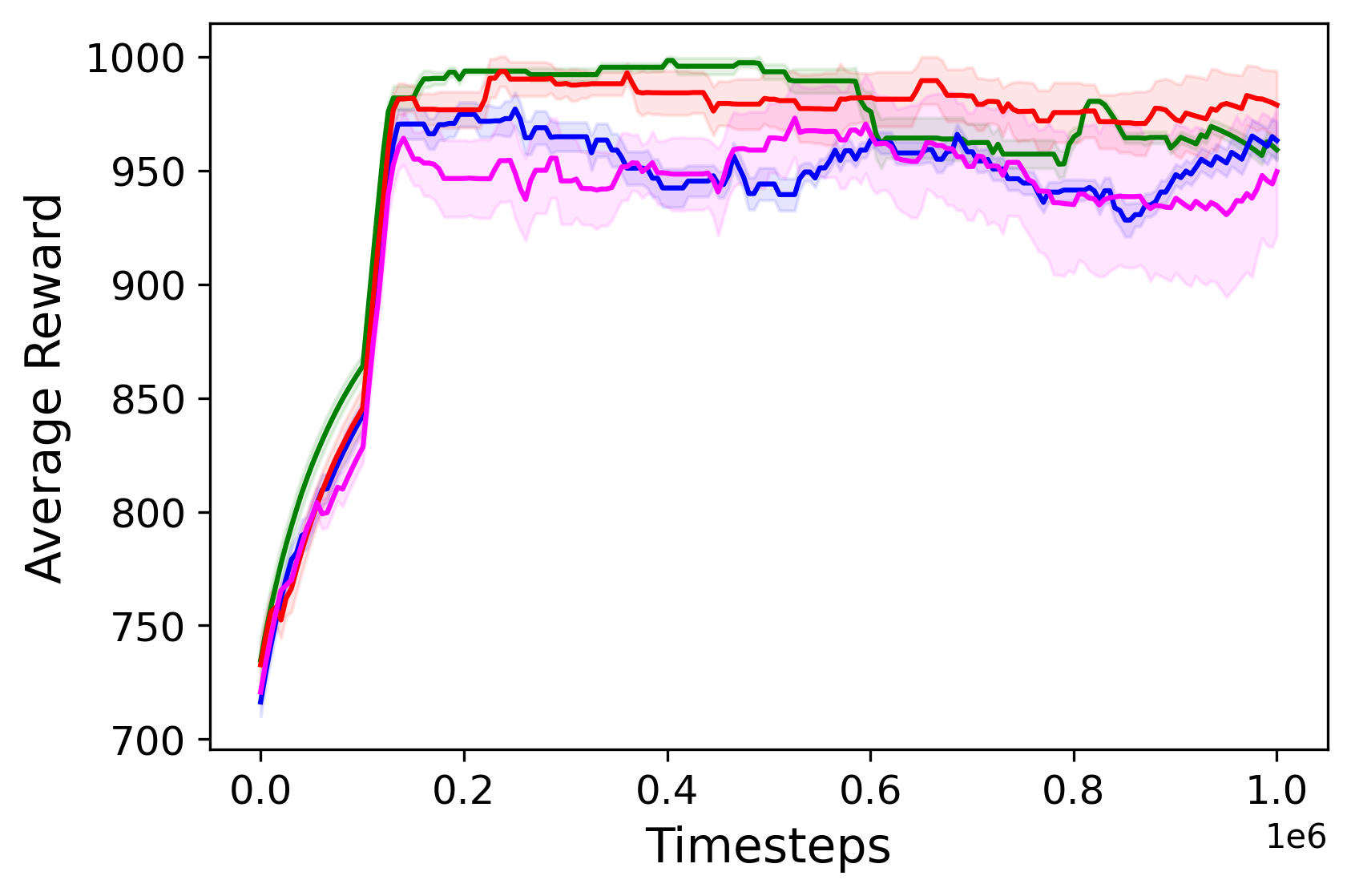}
			\caption{InvertedPendulum-v2}
		\end{subfigure}
		\begin{subfigure}[b]{0.27\textwidth}
			\includegraphics[width=\textwidth]{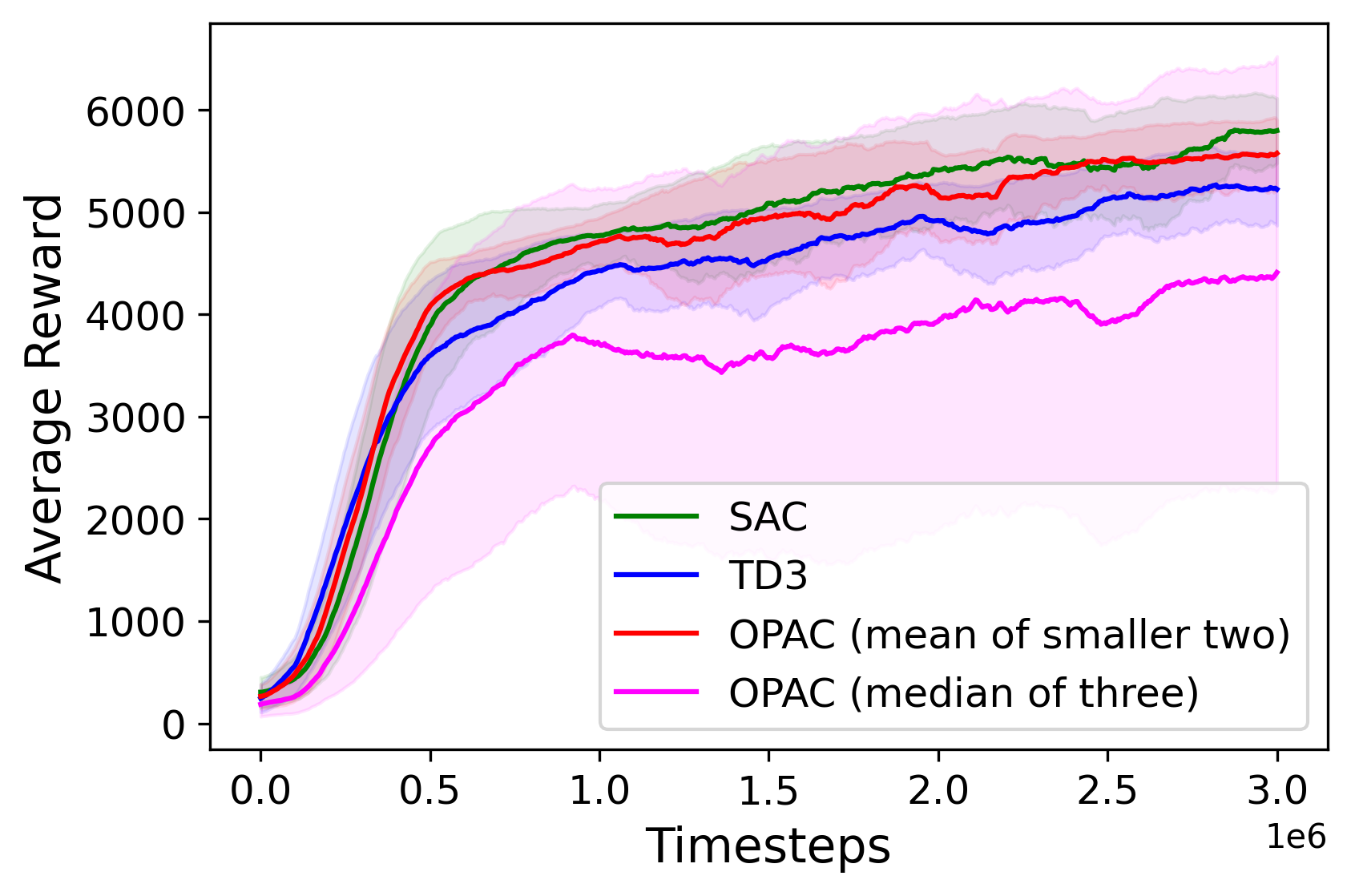}
			\caption{Walker2d-v2}
		\end{subfigure}
		\caption{Learning curves on MuJoCo continuous control environments. %The shaded areas denote one standard deviation of the average evaluation. 
		OPAC outperforms TD3 and SAC in most of the environments. The shaded region corresponds to one standard deviation.}
		\label{fig:Figure1}
	\end{figure*}
	
	\begin{table*}[ht!]
		% \newcolumntype{C}[1]{>{\raggedright\centering\arraybackslash}p{#1}}
		% \newcolumntype{L}[1]{>{\raggedright\arraybackslash}p{#1}}
		\newcolumntype{C}[1]{>{\raggedright\centering\arraybackslash}m{#1}}
		% \newcolumntype{L}[1]{>{\raggedright\arraybackslash}m{#1}}
		\centering
		\begin{tabular}{| l | C{80pt} | C{80pt} | C{80pt} | C{80pt} |}
			\hline
			\textbf{Environment} & \textbf{SAC} & \textbf{TD3} & \textbf{OPAC (mean of the smaller 2 Q-values)} & \textbf{OPAC (median of the 3 Q-values)} \\
			\hline
			\hline
			Ant-v2 & $ 5384.97 \pm 874.90 $ & $ 5547.66 \pm 176.95 $ & $ \bm{7008.67 \pm 120.36} $ & $ 5123.39 \pm 1465.11 $ \\
			\hline
			HalfCheetah-v2 & $ 14469.51 \pm 215.49 $ & $ 11752.63 \pm 292.57 $ & $15185.85 \pm 83.23 $ & $ \bm{15731.22 \pm 73.33} $ \\
			\hline
			Hopper-v2 & $ 3200.85 \pm 373.81 $ & $ 2716.47 \pm 16.74 $ & $ 3375.73 \pm 18.58 $ & $ \bm{3278.37 \pm 572.98} $ \\
			\hline
			Humanoid-v2 & $ 8676.14 \pm 60.60 $ & $ 153.37 \pm 6.45 $ & $ 6925.78 \pm 41.42 $ & $ \bm{9072.19 \pm 95.62} $ \\
			\hline
			InvertedPendulum-v2 & $ \bm{1000.00 \pm 0.00} $ & $ \bm{1000.00 \pm 0.00} $ & $ \bm{1000.00 \pm 0.00} $ & $ \bm{1000.00 \pm 0.00} $ \\
			\hline
			Walker2d-v2 & $ 5969.10 \pm 49.72 $ & $ 5482.32 \pm 58.08 $ & $ 5752.21 \pm 38.10 $ & $ \bm{5228.85 \pm 1518.96 }$ \\
			\hline
		\end{tabular}
		\caption{The maximum average return over 5 trials. The $\pm$ corresponds to one standard deviation. The highest reward in an environment has been boldfaced.}
		\label{tab:Table1}
	\end{table*}
	
	The algorithms were run in Hopper-v2 and InvertedPendulum-v2 for one million time steps whereas in Ant-v2, HalfCheetah-v2, and Walker2d-v2 for three million time steps. Humanoid-v2, the hardest and most challenging environment among all the others, required $10$ million time steps. Figure \ref{fig:Figure1} shows the total average return of evaluation rollouts during training. We train five different instances of each algorithm with the seed values $0$, $200$, $872$, $2359$ and $6574$ and then plot the results by averaging over the five trials. This has been done for the sake of reliability and to make the results reproducible.
	
	The algorithms have been run for $10,000$ time steps with a purely exploratory policy. Policy evaluation is performed after every $5000$ time steps. Each of the evaluation step is performed over $20$ episodes. The evaluation reports the mean of the cumulative reward generated at each of the $20$ episodes without discount and any noise (starting from the start state of the environment as dictated by the seed value).
	The solid curves in Figure \ref{fig:Figure1} corresponds to the mean and the shaded region to one standard deviation of the returns over the five trials. For OPAC, we include both the versions, where we consider mean value of the smaller Q-values (in red) and median value of all the Q-values (in magenta).
	% OPAC (Median) outper
	Table \ref{tab:Table1} shows a comparison between the maximum average reward obtained over the five trials of SAC, TD3 and two variants of OPAC. The curves have been smoothed using simple moving average as needed.
	
	\section{Conclusions}
	In this paper, we presented Opportunistic Actor-Critic (OPAC), an off-policy maximum entropy deep reinforcement learning algorithm that retains the benefits of TD3 and SAC both and also explores better due to the usage of three critics.

	Our theoretical results use the soft policy iteration and automatic entropy adjustment concepts derived in~\cite{DBLP:journals/corr/abs-1812-05905}. These were already shown to converge. We introduced the theory of clipped triple Q-learning and also established its proof of convergence. Combining all these theories, we formulated a practical opportunistic actor-critic algorithm that can be used to train deep neural network policies in continuous state-action spaces. The model is opportunistic in  both action selection and Q-updates. We empirically showed that it equals or exceeds the performance of TD3 and SAC both without any environment-specific hyperparameter tuning.
	Our experiments clearly indicate that OPAC is robust and sample efficient enough
	for easy as well as challenging tasks. It also has lesser variance in its learning curves as shown in Figure~\ref{fig:Figure1} than SAC and TD3. Because of the simplicity of design, OPAC can be included in part to any other actor-critic algorithm.
% 	The name "Opportunistic Actor-Critic" is perfectly justified. It is essentially an actor-critic based off-policy model free deep RL algorithm. 

	%First, we are taking an approach while calculating target Q-values which is neither pessimistic as TD3 and SAC, nor optimistic like in OAC. Second, we delay the policy and target network updates so that the error in soft Q-network is minimized up to a certain extent before introducing a policy update. Third, we are developing our algorithm on the maximum entropy framework which is already proven to improve the quality of policy updates. To summarize, our algorithm is updating its policy and soft Q-function parameters when it gets an appropriate opportunity to do so provided that error in the previous step is kept minimum.
	  \bibliographystyle{plain}
	\bibliography{references}

\end{document}